\documentclass[11pt]{article}
\usepackage{amsmath,amsfonts,bm}
\usepackage[preprint]{acl}
\usepackage{graphicx}
\usepackage[table,xcdraw]{xcolor}
\usepackage{times}
\usepackage{latexsym}
\usepackage{booktabs}
\usepackage[ruled,vlined]{algorithm2e}
\usepackage[T1]{fontenc}

\usepackage[utf8]{inputenc}

\usepackage{microtype}

\usepackage{inconsolata}

\usepackage{graphicx}

\usepackage{eso-pic}
\title{How Reliable are AMR Parser Outputs? \\
Uncertainty Estimation for AMR Parsing}

\author{
Anuj Kumar$^{1,2}$, Shintaro Ozaki$^{2}$, \\ \textbf{Yamuna Prasad}$^{1}$,
\textbf{Katsuhiko Hayashi}$^{2,3}$, \textbf{Hidetaka Kamigaito}$^{2}$, \textbf{Taro Watanabe}$^{2}$ \\
$^{1}$Indian Institute of Technology Jammu, India \\
$^{2}$Nara Institute of Science and Technology (NAIST), Japan \\
$^{3}$The University of Tokyo, Japan \\
\texttt{\{anuj,yamuna.prasad\}@iitjammu.ac.in, katsuhiko-hayashi@g.ecc.u-tokyo.ac.jp} \\
\texttt{\{ozaki.shintaro.ou6,kamigaito.h,taro.watanabe\}@naist.ac.jp}
}

\begin{document}
\setcounter{page}{31144}
\AddToShipoutPictureFG*{%
  \setlength{\unitlength}{1mm}%
  \footnotesize
  \put(0,13){\parbox[t]{\paperwidth}{\centering
    \emph{Proceedings of the 2026 Conference on Empirical Methods in Natural Language Processing}, pages 31144--31159 \\
    October 24-29, 2026 \textcopyright 2026 Association for Computational Linguistics}}%
}
\maketitle
\begin{abstract}
Recent advances have significantly improved the performance of Abstract Meaning Representation (AMR) parsers, yet determining when their outputs can be trusted remains largely unexplored. This work presents the first study of uncertainty estimation for AMR parsers in a black-box setting using top-$k$ beam-search candidates. Uncertainty is estimated from pairwise candidate similarities and evaluated using the Prediction Rejection Ratio. This work characterises the relationship between parsing performance and uncertainty, evaluates the reliability of beam candidate distributions, analyses the effect of evaluation metrics on uncertainty assessment, and identifies parsers that provide the strongest balance between performance and reliability. Experimental results show the value of uncertainty estimation as a complementary dimension for evaluating AMR parsers.
\end{abstract}

\section{Introduction}
Abstract Meaning Representation (AMR) represents sentence meaning as a rooted semantic graph that captures concepts and their relations beyond surface text~\cite{AMR_sembank}. By modelling predicate--argument structure, AMR supports meaning-centric NLP applications including information extraction~\cite{amr_ir}, question answering~\cite{amr_qa}, summarisation~\cite{amr_summerization}, multilingual semantic parsing~\cite{amr_multilingual_parser}, and trustworthy AI~\cite{amr_trust, AMR_parser}.

However, manually constructing AMR graphs is expensive and difficult to scale, motivating the development of automatic AMR parsers that convert text into structured semantic representations. Advances in pretrained language models have substantially improved parsers such as SPRING~\cite{spring}, IBM-Parser~\cite{IBM}, HMG-AMR~\cite{HMG}, MBSE~\cite{MBSE}, and IBM-Neural~\cite{IBM-neural}, making AMR parsing increasingly important for downstream applications where graph quality directly affects reasoning and decision-making.

Despite recent advances in AMR parsing, an important question remains unresolved: how confident should a parser be in its prediction? Recent studies have explored uncertainty estimation for large language models using probabilistic, diversity-based, and graph-based approaches~\cite{lexsim,graph_laplacian,decoding_llm,lm-polygraph}, highlighting the importance of estimating prediction reliability beyond output generation. However, Uncertainty Estimation~(UE) for AMR parsing remains largely unexplored. Existing evaluation protocols rely on graph similarity metrics~\cite{smatch,s2match,sema,semblue,smatchpp,wlk}, which compare predicted graphs against gold annotations and measure aggregate parser performance. Nevertheless, these metrics provide limited insight into the reliability of individual predictions and into whether a parser's confidence aligns with its actual performance. To the best of our knowledge, this work presents the first systematic study of uncertainty estimation for AMR parsing in a black-box setting and addresses the following research questions:

\begin{description}
    \item[RQ1:] Do performance and uncertainty correlate in AMR parsing?
    \item[RQ2:] How reliable are beam candidates?
    \item[RQ3:] Is relying on a single metric enough to judge the uncertainty of AMR parsers?
    \item[RQ4:] Which parser is best in terms of both performance and reliability?
\end{description}

Our experiments\footnote{\url{https://github.com/naist-nlp/UE_Method_AMR_Parser}} evaluate uncertainty estimation across multiple AMR parsers in a black-box setting. For each input, the parser generates a set of beam candidates, from which an uncertainty estimator assigns a scalar uncertainty score; higher scores indicate greater uncertainty. In this work, \textit{Uncertainty Estimation} (UE) refers to the process of assigning these scores, while
\textit{Reliability} refers to how well the scores reflect the actual quality of predictions. The reliability of these uncertainty estimates is evaluated using the Prediction Rejection Ratio~(PRR)~\cite{PRR}, which measures how effectively the uncertainty scores identify low-quality predictions: a reliable estimator should assign higher uncertainty to predictions with lower parsing quality, such that rejecting the most uncertain predictions improves the remaining average quality. Results show a general tendency for higher-performing parsers to produce more reliable uncertainty estimates, although parser architecture and candidate-generation behaviour also influence reliability. Moderate beam sizes yield the most reliable uncertainty estimates, while different evaluation metrics can lead to different conclusions about their reliability. These findings demonstrate that uncertainty estimation provides a complementary perspective beyond parsing accuracy for evaluating AMR parsers.

\section{Problem Definition}

Let $\mathcal{D}=\{(s_i,g_i^*)\}_{i=1}^{n}$ denote a dataset of size $n$, where $s_i$ denotes an input sentence and $g_i^*$ its corresponding gold AMR graph. Given sentence $s_i$, an AMR parser generates $k$ beam candidates
\begin{equation}
\mathcal{C}_i=
\{g_i^{(1)},\ldots,g_i^{(k)}\},
\qquad
g_i^{(1)}=\hat{g}_i,
\end{equation}
where $\mathcal{C}_i$ denotes the set of $k$ candidate AMR graphs, $g_i^{(j)}$ the $j$-th beam prediction, and $\hat g_i$ the top-1 prediction. In the black-box setting, parsing is defined as $P:s_i\mapsto\mathcal C_i$ without access to model probabilities or internal states. The objective is to learn an uncertainty estimator $U$ that maps beam candidates to a scalar uncertainty score,
\begin{equation}
U:\mathcal{C}_i\mapsto u_i,
\qquad
u_i\in\mathbb{R}
\end{equation}
yielding one uncertainty score per input sentence and uncertainty vector $\boldsymbol u=(u_1,\ldots,u_n)\in\mathbb{R}^n$. Prediction quality is computed between the top-1 prediction and gold graph producing quality vector $\boldsymbol q=(q_1,\ldots,q_n)\in\mathbb R^n$. A desirable estimator satisfies $u_i\propto(1-q_i)$ and maximises PRR.

\section{Methodology}
This work estimates uncertainty for AMR parsing in a black-box setting, where model probabilities and internal states are unavailable. The key intuition is that a set of beam candidates with high semantic agreement indicates lower uncertainty, whereas greater disagreement among candidates indicates higher uncertainty. For each input sentence, pairwise similarities among the candidate AMR graphs are computed to construct a similarity matrix (\S\ref{pairwise_similarity}), which is then converted into a scalar uncertainty score (\S\ref{ue}). The quality of these uncertainty estimates is evaluated using the Prediction Rejection Ratio (PRR) (\S\ref{prr}), which measures how effectively the uncertainty scores identify low-quality predictions and whether rejecting the most uncertain predictions improves the quality of the remaining predictions.

\subsection{Pairwise Similarity Matrix}
\label{pairwise_similarity}
For each candidate set $\mathcal{C}_i$, pairwise candidate agreement is
represented by
\begin{equation}
[\mathbf{S}_i]_{jl}
=
F_{\mathrm{sim}}(g_i^{(j)},g_i^{(l)}),
\qquad
\mathbf{S}_i\in\mathbb{R}^{k\times k},
\label{eq:pairwise_similarity}
\end{equation}
where $F_{\mathrm{sim}}$ denotes an AMR similarity function,
$[\mathbf{S}_i]_{jl}\in[0,1]$ measures similarity between beam candidates
$g_i^{(j)}$ and $g_i^{(l)}$, and $[\mathbf{S}_i]_{jj}=1$. This procedure
produces one similarity matrix $\mathbf{S}_i$ per sentence.

\subsection{Uncertainty Estimation}
\label{ue}

Each uncertainty estimator maps beam candidates to an uncertainty score as
$U:\mathcal{C}_i\mapsto\mathbf{S}_i\mapsto u_i$, where $\mathbf{S}_i\in
\mathbb{R}^{k\times k}$ is the pairwise similarity matrix, $u_i\in\mathbb{R}$
is the scalar uncertainty score for sentence $s_i$, and
$\boldsymbol{u}=(u_1,\ldots,u_n)\in\mathbb{R}^n$ is the uncertainty vector.
We use the following estimators:

\paragraph{LexSim}
Following~\citet{lexsim}, lexical uncertainty is estimated from pairwise
ROUGE-L~\cite{rouge} similarity:
\begin{equation}
u_i^{\mathrm{LexSim}}
=
-\frac{1}{k(k-1)}
\sum_{j\neq l}
[\mathbf{R}_i]_{jl}
\end{equation}
where $[\mathbf{R}_i]_{jl}\in[0,1]$ denotes the ROUGE-L similarity between
beam candidates $g_i^{(j)}$ and $g_i^{(l)}$.

\paragraph{Graph-Based Methods}
Following~\citet{graph_laplacian}, $\mathbf{S}_i$ is interpreted as a weighted
graph. The symmetrized adjacency matrix $\mathbf{A}\in\mathbb{R}^{k\times k}$
and degree matrix $\mathbf{D}\in\mathbb{R}^{k\times k}$ are defined as
\begin{equation}
[\mathbf{A}]_{jl}
=
\frac{[\mathbf{S}_i]_{jl}+[\mathbf{S}_i]_{lj}}{2},
\qquad
[\mathbf{D}]_{jj}
=
\sum_l[\mathbf{A}]_{jl}
\end{equation}
and transformed into the normalised Laplacian
\begin{equation}
\mathbf{L}
=
\mathbf{I}-\mathbf{D}^{-1/2}\mathbf{A}\mathbf{D}^{-1/2},
\end{equation}
where $\mathbf{L}\in\mathbb{R}^{k\times k}$ and $\lambda_1,\ldots,\lambda_k$
denote its eigenvalues.
The normalised Laplacian provides a spectral representation of candidate connectivity: the multiplicity of its zero eigenvalue equals the number of connected components in the graph~\cite[Proposition~4]{vonluxburg2007tutorialspectralclustering}. This property motivates the use of the Laplacian eigenvalues and eigenvectors in the following \textbf{EigV} of Equation \ref{eq:engv} and \textbf{Ecc} of Equation \ref{eq:ecc}.

\textbf{EigV} estimates uncertainty from eigenvalue deviations, where larger
deviations $(1-\lambda_j)$ correspond to greater candidate disagreement:
\begin{equation}
u_i^{\mathrm{EigV}}
=
\sum_{j=1}^{k}
\max(0,1-\lambda_j).
\label{eq:engv}
\end{equation}

\textbf{Ecc} estimates uncertainty from candidate spread in the Laplacian
eigenspace. Let $\boldsymbol{v}_1,\ldots,\boldsymbol{v}_k\in\mathbb{R}^{k}$
denote the eigenvectors of $\mathbf{L}$. Each candidate $g_i^{(j)}$ is
embedded as $\boldsymbol{e}_j = (v_{1,j},\ldots,v_{k,j})^\top \in\mathbb{R}^k$
and mean-centred as $\tilde{\boldsymbol{e}}_j = \boldsymbol{e}_j -
\frac{1}{k}\sum_{l=1}^{k}\boldsymbol{e}_l$; uncertainty is the Frobenius
norm of the centred embeddings:
\begin{equation}
u_i^{\mathrm{Ecc}}
=
\sqrt{\sum_{j=1}^{k}\|\tilde{\boldsymbol{e}}_j\|^2}.
\label{eq:ecc}
\end{equation}

\textbf{DegMat} does not rely on the normalized Laplacian and estimates uncertainty from average candidate agreement, where
smaller $\sum_{j,l}[\mathbf{A}]_{jl}$ indicates higher uncertainty:
\begin{equation}
u_i^{\mathrm{DegMat}}
=
1-\frac{\mathrm{tr}(\mathbf{D})}{k^2}
=
1-\frac{1}{k^2}\sum_{j,l}[\mathbf{A}]_{jl}.
\end{equation}

\subsection{Prediction Rejection Evaluation}
\label{prr}

An uncertainty estimator should assign a low uncertainty score to reliable predictions and a high uncertainty score to unreliable predictions. PRR~\cite{PRR} evaluates this behaviour by measuring quality improvement after rejecting uncertain outputs.

Prediction quality is computed between the top-1 prediction $\hat{g}_i$ and
the gold graph $g_i^*$:
\begin{equation}
q_i = \mathcal{F}_{\mathrm{qual}}(\hat{g}_i, g_i^*) \in \mathbb{R}
\label{eq:quality_score}
\end{equation}
yielding quality vector $\boldsymbol{q}=(q_1,\ldots,q_n)\in\mathbb{R}^n$.
Given uncertainty vector $\boldsymbol{u}\in\mathbb{R}^n$, predictions are
sorted by increasing uncertainty $\boldsymbol{\pi}=\mathrm{argsort}
(\boldsymbol{u})\in\mathbb{Z}^n$. At rejection rate $\alpha\in[0,1]$, the
$\lfloor\alpha n\rfloor$ most uncertain predictions are removed, and the remaining mean quality is
\begin{equation}
Q(\alpha)
=
\frac{1}{n-\lfloor\alpha n\rfloor}
\sum_{i=1}^{n-\lfloor\alpha n\rfloor}
q_{\pi_i}
\in\mathbb{R}
\end{equation}
where $\pi_i$ is the $i$-th element of $\boldsymbol{\pi}$. The area under the resulting prediction--rejection curve (AUCPR) is normalised against
oracle and random rejection:
\begin{equation}
\mathrm{PRR}
=
\frac{
\mathrm{AUCPR}_{\mathrm{unc}}
-
\mathrm{AUCPR}_{\mathrm{random}}
}{
\mathrm{AUCPR}_{\mathrm{oracle}}
-
\mathrm{AUCPR}_{\mathrm{random}}
}
\in\mathbb{R}
\end{equation}
where $\mathrm{PRR}=1$ denotes oracle performance, $\mathrm{PRR}=0$ random
rejection, and $\mathrm{PRR}<0$ harmful uncertainty estimates. Further details on the algorithmic framework and AUCPR derivation are
provided in Appendix~\ref{sec:method_framework}.

\section{Experiments and Results}

\subsection{Experimental Setup}
\label{dataset_and_parser}

Experiments are conducted on the AMR~3.0 test set~\cite{AMR_dataset}, comprising $n=1,898$ sentence--graph pairs, using six AMR parsers\footnote{Parsers were selected based on reported performance and code availability. LLM baselines were excluded due to substantially lower parsing performance. Additional details are provided in Appendix~\ref{amr_parser_smatch1}}: SPRING~\cite{spring}, IBM Transition Parser~\cite{IBM}, HMG-AMR-BART~\cite{HMG}, HMG-AMR-T5~\cite{HMG}, MBSE~\cite{MBSE}, and IBM-Neural~\cite{IBM-neural}. For each parser, beam search generates candidate sets with $k\in\{3,5,8,10\}$. Fewer than 1\% of generated beam candidates are invalid across parsers and beam sizes and are assigned zero similarity when computing graph-based similarity measures.

AMR similarity function defined in Equation~\eqref{eq:pairwise_similarity} is measured using Smatch~\cite{smatch}, Smatch++~\cite{smatchpp}, S$^2$Match~\cite{s2match}, SEMA~\cite{sema}, and WLK~\cite{wlk}, while prediction quality defined in Equation~\eqref{eq:quality_score} is evaluated against gold graphs using Smatch++~\cite{smatchpp}, S$^2$Match, and SemBLEU~\cite{semblue}. Additional implementation details, parser checkpoints, similarity metrics,
and hardware configuration are provided in
Appendix~\ref{experiment_setup}.

\begin{table}[t]
\centering
\small
\resizebox{\linewidth}{!}{
\begin{tabular}{lcc}
\toprule
\textbf{Parser} & \textbf{Smatch} & \textbf{Smatch++} \\
\midrule
SPRING ~\cite{spring} & 0.8449 & 0.8438 \\
MBSE ~\cite{MBSE} & 0.8374 & 0.8378 \\
HMG-AMR-BART ~\cite{HMG} & 0.8358 & 0.8380 \\
IBM-Parser ~\cite{IBM} & 0.8128 & 0.8181 \\
IBM-Neural ~\cite{IBM-neural} & 0.8115 & 0.8174 \\
HMG-AMR-T5 ~\cite{HMG} & 0.7693 & 0.7797 \\
\bottomrule
\end{tabular}
}
\caption{Top-1 AMR parsing performance measured using Smatch and Smatch++.}
\label{tab:parser_smatch_smatchpp}
\end{table}

\begin{figure}[t]
    \centering
    \resizebox{\linewidth}{!}{
        \includegraphics{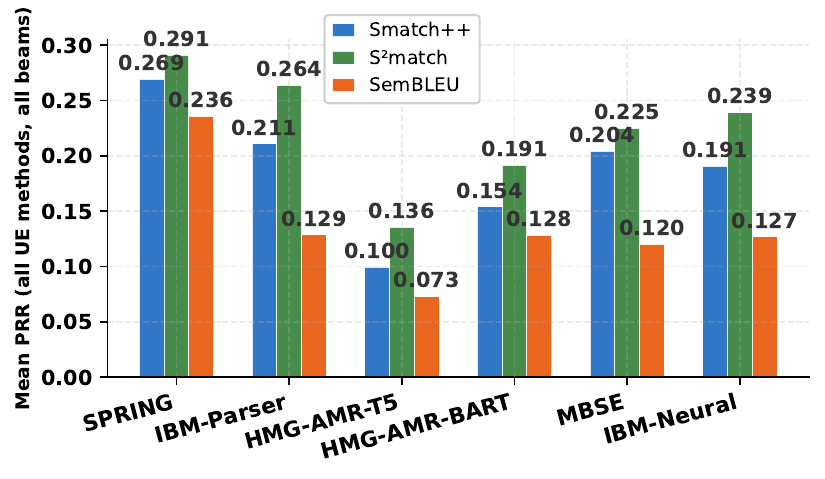}
    }
    \caption{
    Mean PRR for each parser and prediction quality metric, averaged across all beam sizes and uncertainty estimators.
    }
    \label{fig:summary_allbeams}
\end{figure}
\subsection{Results and Analysis}

\paragraph{RQ1: Do performance and uncertainty correlate in AMR parsing?}
Table~\ref{tab:parser_smatch_smatchpp} and Figure~\ref{fig:summary_allbeams} compare parsing performance and PRR score across parsers. Overall, parsers with higher top-1 Smatch scores tend to achieve higher PRR, indicating a positive relationship between parsing accuracy and reliability. However, parsers with comparable Smatch scores exhibit substantially different PRR values; for instance, MBSE and HMG-AMR-BART have mean PRRs of 0.183 and 0.154, respectively, under Smatch++, despite near-identical parsing accuracy, suggesting that candidate-generation behaviour and architectural design also independently influence reliability. Additional results are provided in Appendix~\ref{amr_parser_smatch1} and
Appendix~\ref{amr_parser_smatch2}, where positive correlations are observed
consistently across evaluation metrics. Quantitatively, the Spearman
correlation between parsing performance and mean PRR is $\rho=0.388$ for
Smatch++, $\rho=0.317$ for S$^2$Match, and $\rho=0.360$ for SemBLEU, with
the strongest correlation observed under Smatch++. These results further
support the positive relationship between parsing performance and
uncertainty-estimation reliability.

\begin{figure}[t]
    \centering
    \includegraphics[width=\linewidth]{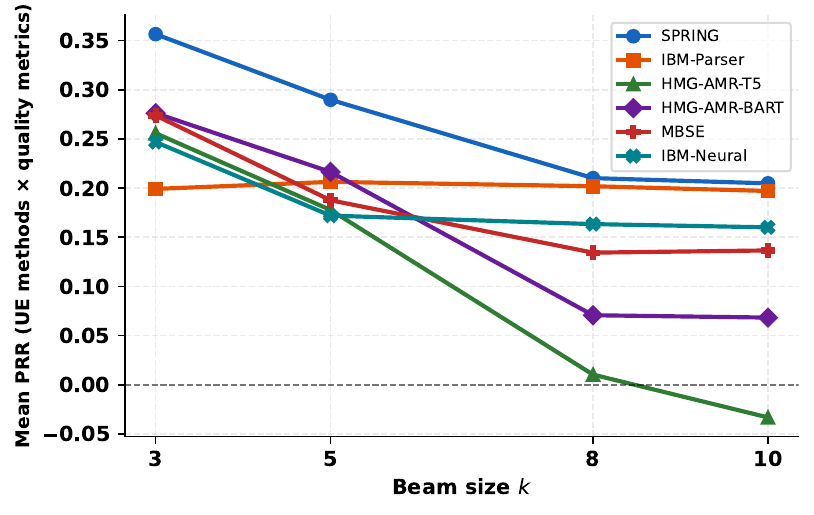}
    \caption{
    Mean PRR averaged across uncertainty estimation methods and prediction quality metrics for beam sizes $k\in\{3,5,8,10\}$.
    }
    \label{fig:beam_effect}
\end{figure}

\paragraph{RQ2: How reliable are beam candidates?}
Beam candidates provide meaningful uncertainty estimates, as shown in Figure~\ref{fig:beam_effect}. However, PRR declines across most parsers as beam size increases, suggesting that larger beams introduce a more diverse set of candidates, weakening the relationship between uncertainty and prediction quality. The rate of degradation varies substantially: HMG-AMR-T5 falls below zero at $k=10$, indicating that the uncertainty estimates are actively harmful, whereas transition-based parsers remain comparatively stable, potentially because their architecture explores structured action sequences rather than relying solely on graph linearisation. Overall, moderate beam sizes provide the strongest trade-off between candidate diversity and uncertainty quality. Additional analysis across beam sizes and uncertainty estimators is provided in Appendix~\ref{amr_parser_RQ2}, where \textsc{LexSim} is the only estimator whose PRR consistently increases with beam size, whereas \textsc{Ecc} exhibits the steepest degradation. These results show that increasing the beam size does not affect all uncertainty estimators equally, as their ability to capture candidate agreement varies.

\begin{figure}[!t]
    \centering
    \resizebox{\linewidth}{!}{
    \includegraphics{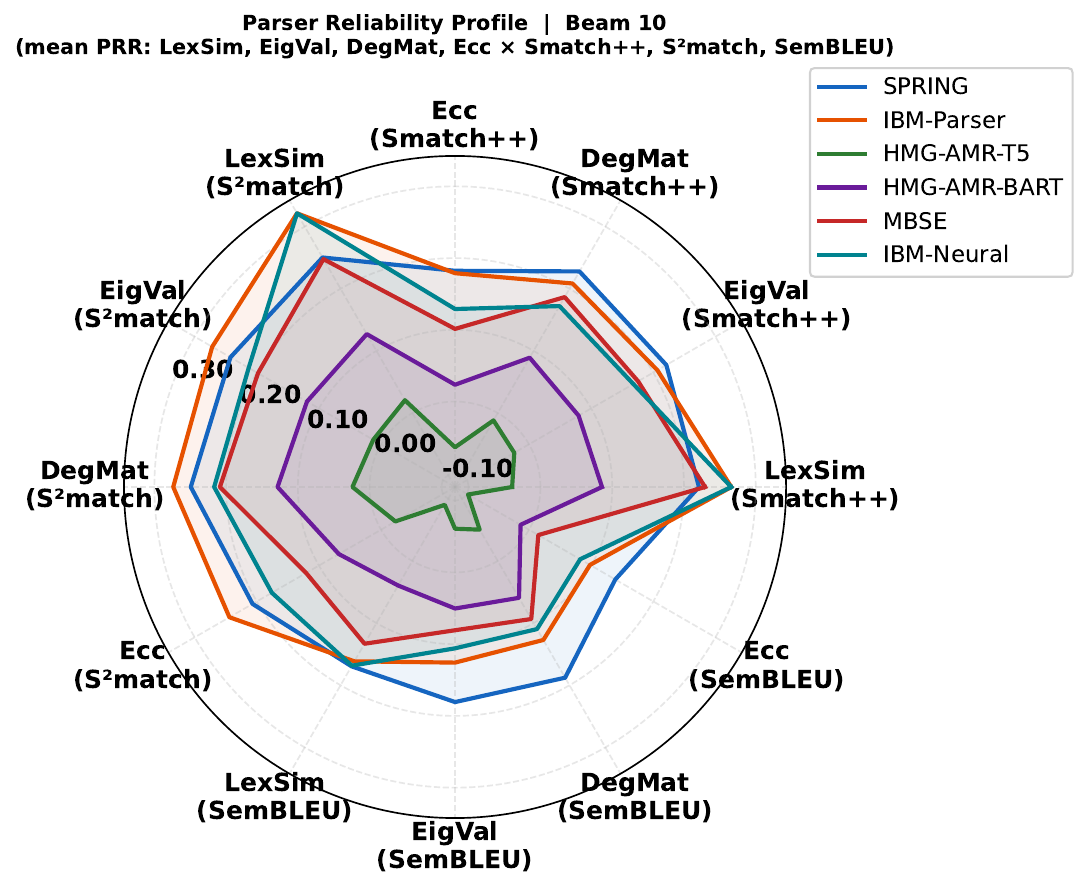}
    }
    \caption{
    Each value shows the mean PRR averaged across uncertainty estimation methods and prediction quality metrics.
    }
    \label{fig:beam10_parser_profile}
\end{figure}

\paragraph{RQ3: Is relying on a single metric enough to judge the uncertainty of AMR parsers?}
No, the uncertainty of AMR parsers cannot be judged by using a single evaluation metric. As illustrated in Figure~\ref{fig:beam10_parser_profile}, PRR varies across metrics, with transition-based parsers achieving stronger PRR under S$^2$Match than Smatch++, indicating that uncertainty assessment depends on the notion of graph similarity being measured. Additional analysis is provided in Appendix~\ref{amr_parser_RQ3}, where graph-based estimators show strong mutual agreement under Smatch++, while \textsc{LexSim} shows near-zero or negative correlation with most graph-based methods. Under SemBLEU, estimator agreement becomes more uniform, indicating that the relationship between uncertainty estimators also changes with the evaluation metric. These observations reinforce that reliability is jointly determined by the uncertainty estimator and the metric used to assess prediction quality, making multi-metric evaluation important for a robust assessment.

\paragraph{RQ4: Which parser is best in terms of both performance and reliability?} SPRING~\cite{spring} achieves the strongest overall balance, obtaining the highest mean PRR (0.265) while maintaining positive uncertainty estimates across all beam sizes and evaluation metrics as shown in Figure~\ref{fig:summary_allbeams} and \ref{fig:beam_effect}. Its seq2seq formulation with graph-isomorphic linearisation reduces reliance on heuristic processing, preserving semantically coherent candidate sets as beam diversity increases. Additional parser reliability results across beam sizes are provided in Appendix~\ref{amr_parser_RQ4_1}, where IBM-Parser remains the most stable across all metrics. At the same time, HMG-AMR-T5 exhibits the steepest degradation, approaching zero or negative PRR at $k=8$ and $k=10$. Detailed estimation results at $k=10$ are provided in Appendix~\ref{amr_parser_RQ4_2}, where \textsc{DegMat-SEMA} achieves the strongest PRR for SPRING and HMG variants, while \textsc{LexSim} outperforms graph-based methods for IBM-Parser and IBM-Neural. These results highlight that parser selection involves a trade-off between overall parsing performance and the stability of uncertainty estimation across beam sizes. While SPRING provides the strongest overall balance, IBM-Parser may be preferable when stable uncertainty estimation across different beam sizes is prioritised.

\subsection{Additional Analysis}

To further investigate factors affecting uncertainty estimation and reliability, the analysis examines sentence length and agreement between uncertainty estimators. Short sentences ($\leq 9$ tokens) receive substantially higher mean uncertainty scores than medium- and long-length sentences; however, PRR indicates that uncertainty estimates are most reliable for medium-length sentences, followed by long and short sentences. The analysis further shows strong agreement among graph-based estimators. Additional analysis
is provided in Appendix~\ref{amr_parser_RQ2} and \ref{amr_parser_RQ4_2} showing that parser reliability varies substantially with beam size: some parsers remain stable as the beam expands, whereas others exhibit pronounced degradation. At larger beam sizes, the most reliable UE also varies across parsers, indicating that no single estimator consistently provides the strongest reliability.

\section{Conclusion}
This work presents UE for AMR parsers in a black-box setting. Across multiple AMR parsers, beam sizes, and evaluation metrics, it is observed that more accurate parsers generally yield more reliable uncertainty estimates. At the same time, parser reliability also varies with beam size and parser architecture. The results further show that the reliability of UE depends on the evaluation metric and that moderate beam sizes tend to yield more reliable UE than larger beams. Importantly, the experiments demonstrate that parsing performance and uncertainty estimation with reliability provide complementary information for comparing AMR parsers. Parsing accuracy measures the quality of the predicted AMRs, whereas PRR evaluates how effectively the estimated uncertainty identifies potentially low-quality predictions. Since parsers with similar parsing accuracy can exhibit substantially different PRR, considering both performance and reliability provides a more comprehensive assessment of parser behaviour. Overall, the findings support the use of both conventional accuracy-based evaluation and uncertainty-based reliability evaluation when comparing AMR parsers.

\section*{Limitations}

This work studies uncertainty estimation for AMR parsing under a black-box setting using beam-search candidates as the sole source of uncertainty. Although beam decoding is adopted by most publicly available AMR parsers and therefore provides a consistent evaluation setting across models, the behaviour of uncertainty under alternative decoding strategies (e.g., sampling-based or diverse decoding methods) remains unexplored and may yield different candidate distributions and reliability characteristics.

A second limitation concerns evaluation. Uncertainty quality in this study is measured using the PRR alongside existing AMR quality metrics. However, current AMR evaluation metrics capture distinct aspects of graph quality and can yield different conclusions about uncertainty, as observed in our experiments. This suggests that uncertainty-aware evaluation for AMR remains an open problem and that future work may benefit from developing reliability-oriented evaluation protocols that complement conventional parsing metrics.

Finally, this study focuses on post-hoc uncertainty estimation without modifying parser architectures or training objectives. Integrating uncertainty directly into AMR parser training and decoding may further improve reliability.

\section*{Acknowledgment}
The authors would like to thank the Japan Science and Technology Agency (JST) for supporting the research activity for this project at the Nara Institute of Science and Technology (NAIST) through the LOTUS Programme (India-Japan Circulation of Talented Youths in Science Programme).

\section*{Ethical Considerations}

AI-assisted tools were used to support parts of the research workflow. ChatGPT\footnote{\url{https://chat.openai.com/}} was used for writing support and language refinement, Claude Code\footnote{\url{https://www.anthropic.com/claude-code}} was used to accelerate research workflows, and Grammarly\footnote{\url{https://www.grammarly.com/}} was used for grammar correction and proofreading. All experiments, analyses, research decisions, and final manuscript content were conducted and verified by the authors.

\bibliography{custom}

\appendix

\section{Algorithm}
\label{sec:method_framework}

\begin{algorithm}[!h]
\small
\caption{Black-box Uncertainty Estimation for AMR Parsing}
\label{alg:amr_ue}
\KwIn{
Dataset $\mathcal{D}=\{(s_i,g_i^*)\}_{i=1}^{n}$,
parser $P$, beam size $k$
}
\KwOut{
PRR $\mathrm{PRR}\in\mathbb{R}$
}

\ForEach{$(s_i,g_i^*)\in\mathcal{D}$}{
Generate $k$ beam candidates via beam search\;
\[
\mathcal{C}_i=P(s_i)=\{g_i^{(1)},\ldots,g_i^{(k)}\}
\]

Compute pairwise candidate similarity matrix\;
\[
\mathbf{S}_i=F_{\mathrm{sim}}(\mathcal{C}_i)\in\mathbb{R}^{k\times k},
\qquad [\mathbf{S}_i]_{jj}=1,
\]
\[
[\mathbf{S}_i]_{jl}\in[0,1]
\]

UE methods\;
\[
\begin{aligned}
u_i^{\mathrm{LexSim}}
&=
1-\frac{1}{k(k-1)}
\sum_{j\neq l}
\mathrm{ROUGE}(g_i^{(j)},g_i^{(l)}),
\\
u_i^{\mathrm{DegMat}}
&=
1-
\frac{2}{k(k-1)}
\sum_{j<l}[\mathbf{S}_i]_{jl},
\\
u_i^{\mathrm{EigV}}
&=
\frac{1}{k}\sum_{r=1}^{k}(\lambda_r-1)^2,
\\
u_i^{\mathrm{Ecc}}
&=
\left(
\sum_{j=1}^{k}
\left\|
\mathbf{e}_j-\frac{1}{k}\sum_{l=1}^{k}\mathbf{e}_l
\right\|_2^2
\right)^{1/2}.
\end{aligned}
\]
Higher $u_i$ indicates greater uncertainty\;

Compute prediction quality against the gold graph\;
\[
q_i=
\mathcal{F}_{\mathrm{qual}}(g_i^{(1)},g_i^*)
\in\mathbb{R}
\]
}

Construct uncertainty and quality vectors\;
\[
\boldsymbol{u}=(u_1,\ldots,u_n)\in\mathbb{R}^n,
\qquad
\boldsymbol{q}=(q_1,\ldots,q_n)\in\mathbb{R}^n
\]

Sort by increasing uncertainty ($\pi_1$ = least uncertain)\;
\[
\boldsymbol{\pi}
=
\mathrm{argsort}(\boldsymbol{u})
\in\mathbb{Z}^n
\]

At rejection rate $\alpha$, remove $\lfloor\alpha n\rfloor$ most uncertain
predictions and compute remaining mean quality\;
\[
Q(\alpha)
=
\frac{1}{n-\lfloor\alpha n\rfloor}
\sum_{r=1}^{n-\lfloor\alpha n\rfloor}
q_{\pi_r}
\]

Compute area under the prediction--rejection curve\;
\[
\mathrm{AUCPR}
=
\int_0^1 Q(\alpha)\,\mathrm{d}\alpha
\]

Normalise against oracle and random baselines\;
\[
\mathrm{PRR}
=
\frac{
\mathrm{AUCPR}_{\mathrm{unc}}
-
\mathrm{AUCPR}_{\mathrm{random}}
}{
\mathrm{AUCPR}_{\mathrm{oracle}}
-
\mathrm{AUCPR}_{\mathrm{random}}
}
\]

\Return $\mathrm{PRR}$
\end{algorithm}

\subsection{Algorithm}
Algorithm~\ref{alg:amr_ue} summarises the complete procedure.

\subsection{Computational Complexity}

The dominant cost of the proposed framework is pairwise similarity computation, which requires $\mathcal{O}(k^2)$ graph comparisons per sentence. Alignment-based metrics (Smatch, Smatch++, S$^2$Match) scale as $\mathcal{O}(|V|^2 r)$ per comparison, where $|V|$ is the number of graph nodes and $r$ the number of alignment restarts; SEMA scales as $\mathcal{O}(|V|+|E|)$; and WLK scales as $\mathcal{O}(|V||E|h)$, where $|E|$ is the number of edges and $h$ the number of Weisfeiler--Lehman iterations. The overall runtime across $P$ parsers, $B$ beam configurations, and $N$ sentences is therefore $\mathcal{O}(P \cdot B \cdot N \cdot k^2 \cdot C_{\mathrm{sim}})$, where $C_{\mathrm{sim}}$ denotes the per-comparison cost of the chosen metric. Uncertainty estimation adds negligible overhead: \textsc{DegMat} and \textsc{Ecc} operate in $\mathcal{O}(Nk^2)$, \textsc{EigV} in $\mathcal{O}(Nk^3)$, and PRR evaluation in $\mathcal{O}(N\log N)$ per estimator--metric pair. To avoid redundant computation, all pairwise similarity matrices are cached as compressed \texttt{.npz} arrays after the first computation.

\section{Experiment Setup}
\label{experiment_setup}

This Appendix provides implementation details required to reproduce all experiments.

\subsection{Dataset}
\label{appendix_dataset}
Experiments are conducted on the AMR~3.0 (\url{https://catalog.ldc.upenn.edu/LDC2020T02}) test set containing 1,898 sentence--graph pairs. Each example consists of an input sentence and its corresponding gold AMR graph represented in PENMAN notation. All uncertainty estimation experiments are performed on parser outputs generated from this test split.

\subsection{AMR Parsers}
\label{appendix_amr_parser}

Six publicly available AMR parsers are evaluated in this study. All experiments use official implementations and publicly released pretrained checkpoints without modifying model architectures or retraining.

\begin{itemize}
\item SPRING~\cite{spring}
(\url{https://github.com/SapienzaNLP/spring})

\item IBM-Parser~\cite{IBM}
(\url{https://github.com/IBM/transition-amr-parser})

\item MBSE~\cite{MBSE} using checkpoint:
amr3joint\_ontowiki2\_g2g-structured-bart-large-seed44, loaded through the IBM implementation (\url{https://github.com/IBM/transition-amr-parser}).

\item IBM-Neural~\cite{IBM-neural} using checkpoint:
amr3.0-structured-bart-large-neur-al-seed42, loaded through the IBM implementation (\url{https://github.com/IBM/transition-amr-parser}).

\item HMG-AMR-T5~\cite{HMG}
(\url{https://github.com/YIKMAT/HMG-AMR/tree/main/HMG-AMR(T5)})

\item HMG-AMR-BART~\cite{HMG}
(\url{https://github.com/YIKMAT/HMG-AMR/tree/main/HMG-AMR(Bart)})
\end{itemize}

For reference, two instruction-tuned LLM baselines (Qwen3-32B and Ministral-14B) were additionally evaluated using direct AMR generation without task-specific training using the transformer library~\cite{transformers}. Due to substantially lower performance on AMR~3.0 as shown in Table~\ref{tab:parser_smatch} and the absence of dedicated AMR parsing capability, these models were excluded from uncertainty estimation experiments.

\subsection{Similarity Construction and Evaluation}
\label{appendix_evaluation_matrix}

Pairwise candidate similarity matrices $\mathbf{S}_i \in \mathbb{R}^{k\times k}$
are constructed using four AMR-specific similarity metrics:

\begin{itemize}
\item Smatch~\cite{smatch}: \url{https://github.com/flipz357/amr-metric-suite}
\item S$^2$Match~\cite{s2match}: \url{https://github.com/flipz357/amr-metric-suite}
\item SEMA~\cite{sema}: \url{https://github.com/flipz357/amr-metric-suite}
\item WLK~\cite{wlk}: \url{https://github.com/flipz357/weisfeiler-leman-amr-metrics}
\end{itemize}

Prediction quality for PRR evaluation is measured using three metrics:

\begin{itemize}
\item Smatch++~\cite{smatchpp}: \url{https://github.com/flipz357/smatchpp.git}
\item S$^2$Match~\cite{s2match}: \url{https://github.com/flipz357/amr-metric-suite}
\item SemBLEU~\cite{semblue}: \url{https://github.com/freesunshine0316/sembleu}
\end{itemize}

\subsection{Uncertainty Estimation and Evaluation}
\label{appendix_ue_methods}

Uncertainty is estimated using one lexical and three graph-based approaches:

\begin{itemize}
\item \textsc{LexSim}~\cite{lexsim}: estimates uncertainty from mean pairwise ROUGE-L similarity across beam candidates; lower lexical overlap indicates greater candidate diversity and higher uncertainty.

\item \textsc{EigV}~\cite{graph_laplacian}: constructs a similarity graph over candidates and measures uncertainty from the eigenvalues of the normalised Laplacian; larger eigenvalue deviations from one indicate weaker graph connectivity and greater candidate disagreement.

\item \textsc{DegMat}~\cite{graph_laplacian}: measures uncertainty as the complement of the normalised total pairwise similarity; a lower sum of edge weights in the similarity graph indicates higher uncertainty.

\item \textsc{Ecc}~\cite{graph_laplacian}: embeds each candidate in the Laplacian eigenspace and measures uncertainty as the spread of embeddings around their centroid; a larger Frobenius norm indicates greater candidate dispersion and higher uncertainty.
\end{itemize}

Prediction quality for PRR evaluation is measured using
Smatch++~\cite{smatchpp}, S$^2$Match~\cite{s2match}, and
SemBLEU~\cite{semblue}. Uncertainty quality is evaluated using the
PRR~\cite{PRR}.

\subsection{Hardware and Runtime}

All uncertainty estimation and PRR calculations were performed in a CPU-only environment without GPU acceleration. Each job was allocated 4 CPU cores and
64\,GB of RAM.

Table~\ref{tab:runtime} reports the observed runtime measured from execution logs. Generating beam candidates and computing uncertainty metrics required approximately 25 hours per parser across all beam sizes ($k\in\{3,5,8,10\}$). Processing a single beam configuration required approximately 6 hours. Running the complete evaluation pipeline across all six parsers took approximately 150 hours of wall-clock time.

\begin{table}[t]
\centering
\small
\begin{tabular}{ll}
\toprule
\textbf{Configuration} & \textbf{Runtime} \\
\midrule
Per parser (all beams) &
$\sim$89,000--90,000 s ($\approx$25 h) \\

Per parser per beam &
$\sim$22,000 s ($\approx$6 h) \\

Full experiment (6 parsers) &
$>$150 h wall-clock \\
\bottomrule
\end{tabular}
\caption{
Hardware allocation and observed execution time were measured from job logs.
}
\label{tab:runtime}
\end{table}

\section{Additional Results}
\label{additional_result}

This section presents additional analyses supporting the claims in the main paper, including parser-level performance comparisons, the correlation between AMR parsing quality and uncertainty estimation, and the stability of uncertainty estimators under varying beam sizes.

\begin{table}[t]
\centering
\small
\resizebox{\linewidth}{!}{
\begin{tabular}{lc}
\toprule
\textbf{Parser} & \textbf{Smatch} \\
\midrule
SPRING ~\cite{spring} & 0.8449 \\
MBSE ~\cite{MBSE} & 0.8374 \\
HMG-AMR-BART ~\cite{HMG} & 0.8358 \\
IBM-Parser ~\cite{IBM} & 0.8128 \\
IBM-Neural ~\cite{IBM-neural} & 0.8115 \\
HMG-AMR-T5 ~\cite{HMG} & 0.7693 \\
\midrule
Qwen3-32B ~\cite{qwen} & 0.1214 \\
Ministral-14B ~\cite{ministral} & 0.0897 \\
\bottomrule
\end{tabular}
}
\caption{Top-1 AMR parsing performance measured using Smatch.}
\label{tab:parser_smatch}
\end{table}

\begin{figure*}[t]
\centering
\resizebox{\textwidth}{!}{
\includegraphics{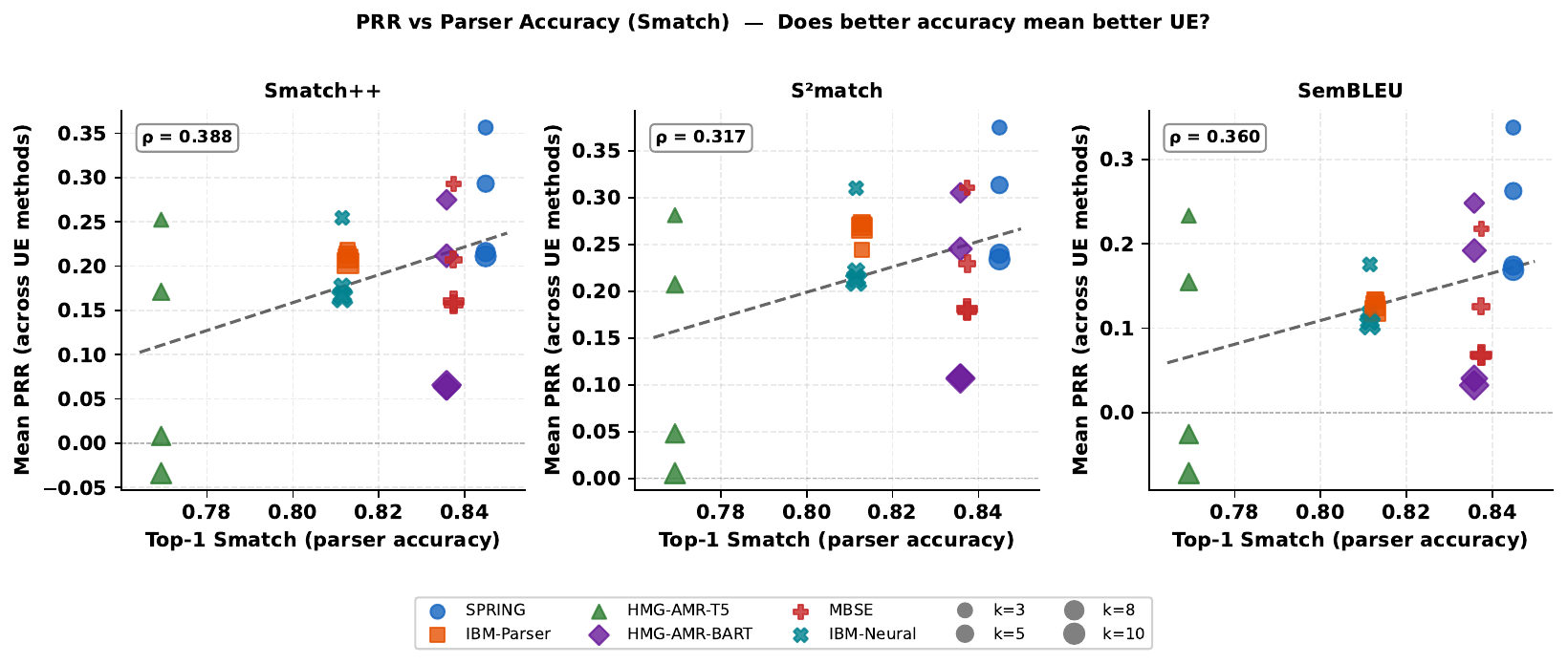}
}
\caption{Relationship between parser accuracy (Smatch) and uncertainty quality (PRR) across beam sizes and evaluation metrics.}
\label{fig:prr_vs_smatch}
\end{figure*}

\subsection{AMR Parsing Performance}
\label{amr_parser_smatch1}

Table~\ref{tab:parser_smatch} reports parser Smatch score performance on the gold dataset reproduced using the publicly released implementations and default hyperparameter settings provided in the original papers. To ensure consistent comparison across parsers, all results are evaluated before BLINKification. Under this setting, SPRING achieves the highest parsing accuracy (0.8449), followed by MBSE (0.8374) and HMG-AMR-BART (0.8358), indicating that recent encoder--decoder approaches remain highly competitive for AMR generation. IBM-Parser and IBM-Neural obtain comparable performance around 0.81 Smatch, whereas HMG-AMR-T5 shows lower accuracy relative to other specialised AMR parsers. For reference, two instruction-tuned LLM baselines (Qwen3-32B and Ministral-14B) are additionally evaluated using direct AMR generation without task-specific training and exhibit substantially lower performance, highlighting the difficulty of generating valid AMR structures through general-purpose language modelling alone. Due to their poor performance on the AMR~3.0 test set, these models are excluded from subsequent analysis.

\begin{figure*}[t]
\centering
\resizebox{\textwidth}{!}{
\includegraphics{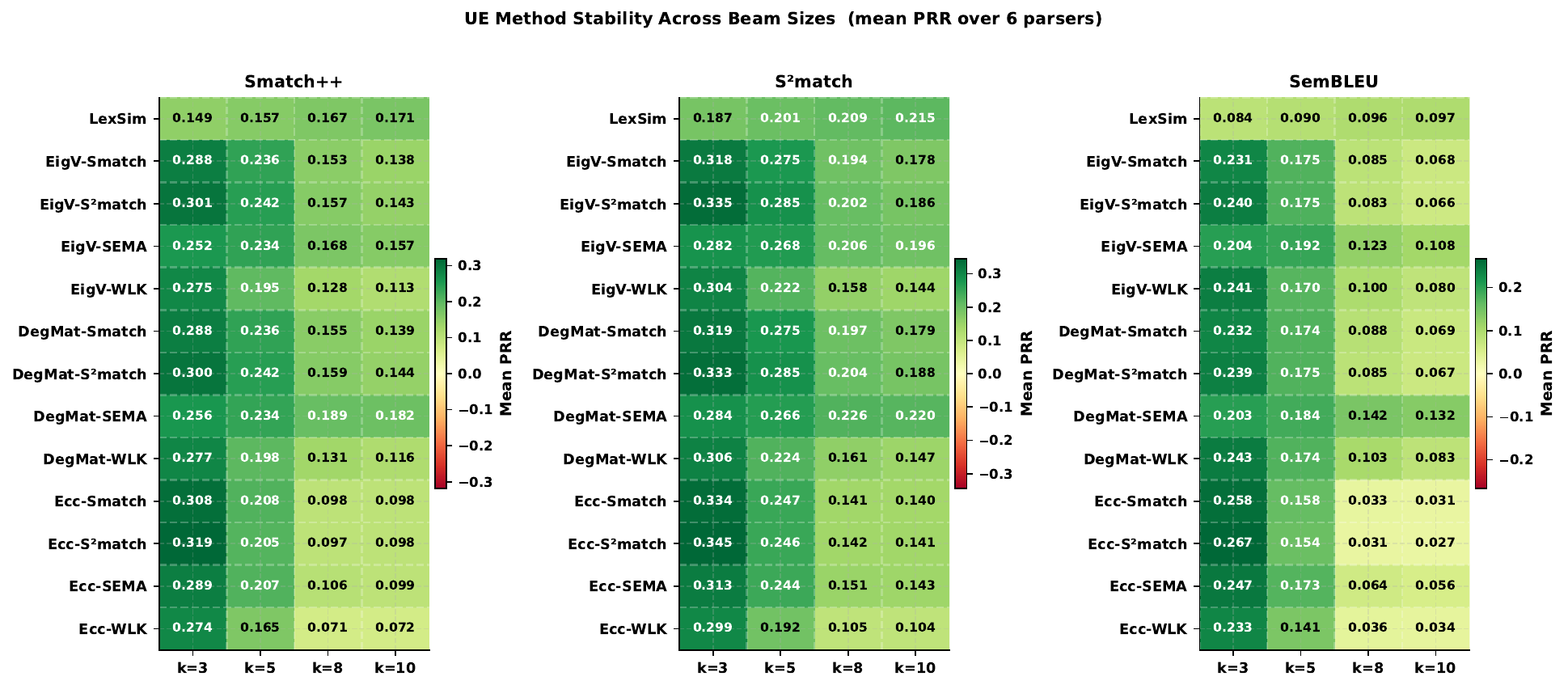}
}
\caption{Mean PRR of uncertainty estimators across beam sizes aggregated over all parsers. Stable estimators exhibit less variation in performance across beam sizes.}
\label{fig:ue_stability}
\end{figure*}

\begin{figure*}[t]
\centering
\resizebox{\textwidth}{!}{
\includegraphics{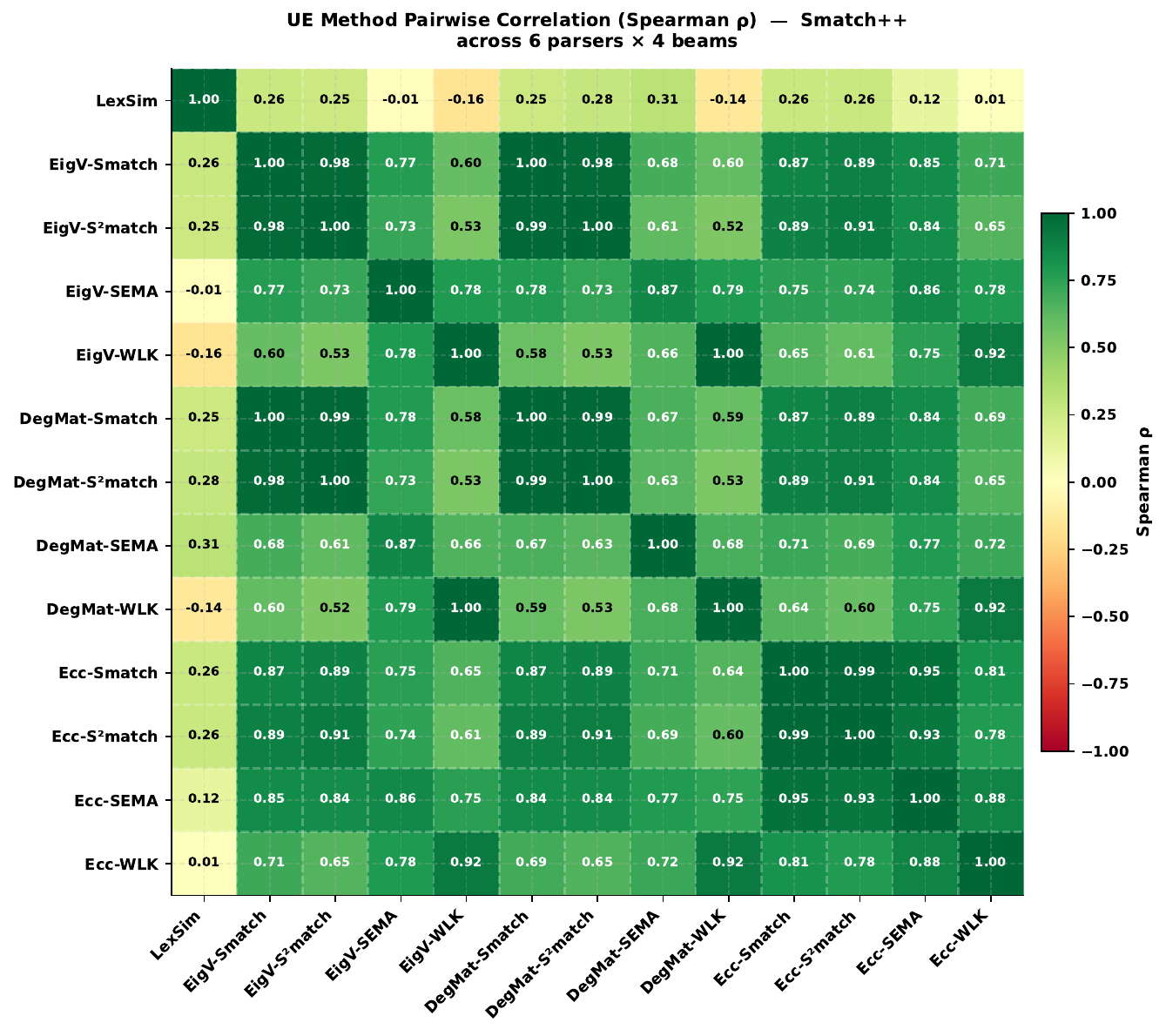}
}
\caption{
Pairwise Spearman correlation between uncertainty estimators evaluated using Smatch++. Graph-based estimators exhibit strong agreement, while \textsc{LexSim} captures a distinct uncertainty signal.
}
\label{fig:ue_corr_spp}
\end{figure*}

\begin{table*}[t]
\resizebox{\textwidth}{!}{%
\begin{tabular}{lcccccccccccc}
\rowcolor[HTML]{1A3A5C}
{\color[HTML]{FFFFFF} \textbf{}} &
  \multicolumn{2}{c}{\cellcolor[HTML]{1A3A5C}{\color[HTML]{FFFFFF} \textbf{SPRING}}} &
  \multicolumn{2}{c}{\cellcolor[HTML]{1A3A5C}{\color[HTML]{FFFFFF} \textbf{IBM -Parser}}} &
  \multicolumn{2}{c}{\cellcolor[HTML]{1A3A5C}{\color[HTML]{FFFFFF} \textbf{HMG-T5}}} &
  \multicolumn{2}{c}{\cellcolor[HTML]{1A3A5C}{\color[HTML]{FFFFFF} \textbf{HMG-BART}}} &
  \multicolumn{2}{c}{\cellcolor[HTML]{1A3A5C}{\color[HTML]{FFFFFF} \textbf{MBSE}}} &
  \multicolumn{2}{c}{\cellcolor[HTML]{1A3A5C}{\color[HTML]{FFFFFF} \textbf{IBM-Neural}}} \\
\rowcolor[HTML]{2E6DA4}
\cellcolor[HTML]{BDD7EE}{\color[HTML]{1A3A5C} \textbf{UE Method}} &
  {\color[HTML]{FFFFFF} \textbf{S++}} &
  {\color[HTML]{FFFFFF} \textbf{SB}} &
  {\color[HTML]{FFFFFF} \textbf{S++}} &
  {\color[HTML]{FFFFFF} \textbf{SB}} &
  {\color[HTML]{FFFFFF} \textbf{S++}} &
  {\color[HTML]{FFFFFF} \textbf{SB}} &
  {\color[HTML]{FFFFFF} \textbf{S++}} &
  {\color[HTML]{FFFFFF} \textbf{SB}} &
  {\color[HTML]{FFFFFF} \textbf{S++}} &
  {\color[HTML]{FFFFFF} \textbf{SB}} &
  {\color[HTML]{FFFFFF} \textbf{S++}} &
  {\color[HTML]{FFFFFF} \textbf{SB}} \\
\rowcolor[HTML]{EBF3FB}
LexSim &
  0.2209 &
  0.1695 &
  \cellcolor[HTML]{C6EFCE}{\color[HTML]{276221} \textbf{0.2660}} &
  \cellcolor[HTML]{C6EFCE}{\color[HTML]{276221} \textbf{0.1615}} &
  -0.0395 &
  -0.0904 &
  0.0858 &
  0.0395 &
  \cellcolor[HTML]{C6EFCE}{\color[HTML]{276221} \textbf{0.2294}} &
  0.1332 &
  \cellcolor[HTML]{C6EFCE}{\color[HTML]{276221} \textbf{0.2659}} &
  \cellcolor[HTML]{C6EFCE}{\color[HTML]{276221} \textbf{0.1685}} \\
\rowcolor[HTML]{FFFFFF}
EigV-Smatch    & 0.1956 & 0.1421 & 0.2218 & 0.1251 & -0.0838 & -0.1199 & 0.0756 & 0.0359  & 0.1871 & 0.0802 & 0.2325 & 0.1431 \\
\rowcolor[HTML]{EBF3FB}
EigV-S2match   & 0.1912 & 0.1322 & 0.2319 & 0.1232 & -0.0794 & -0.1176 & 0.0751 & 0.0330  & 0.1931 & 0.0752 & 0.2448 & 0.1473 \\
\rowcolor[HTML]{FFFFFF}
EigV-SEMA      & 0.2504 & 0.2254 & 0.1919 & 0.1302 & 0.0334  & -0.0060 & 0.1080 & 0.0811  & 0.1666 & 0.0859 & 0.1945 & 0.1328 \\
\rowcolor[HTML]{EBF3FB}
EigV-WLK       & 0.2469 & 0.2231 & 0.1807 & 0.1243 & 0.0348  & -0.0002 & 0.0605 & 0.0512  & 0.1555 & 0.0811 & 0.0000 & 0.0000 \\
\rowcolor[HTML]{FFFFFF}
DegMat-Smatch  & 0.1971 & 0.1438 & 0.2220 & 0.1250 & -0.0826 & -0.1183 & 0.0776 & 0.0380  & 0.1878 & 0.0807 & 0.2328 & 0.1434 \\
\rowcolor[HTML]{EBF3FB}
DegMat-S2match & 0.1925 & 0.1336 & 0.2321 & 0.1234 & -0.0755 & -0.1141 & 0.0772 & 0.0351  & 0.1942 & 0.0763 & 0.2452 & 0.1477 \\
\rowcolor[HTML]{C6EFCE}
\cellcolor[HTML]{FFFFFF}DegMat-SEMA &
  {\color[HTML]{276221} \textbf{0.2709}} &
  {\color[HTML]{276221} \textbf{0.2465}} &
  \cellcolor[HTML]{FFFFFF}0.1984 &
  \cellcolor[HTML]{FFFFFF}0.1353 &
  {\color[HTML]{276221} \textbf{0.0685}} &
  {\color[HTML]{276221} \textbf{0.0232}} &
  {\color[HTML]{276221} \textbf{0.1369}} &
  {\color[HTML]{276221} \textbf{0.1091}} &
  \cellcolor[HTML]{FFFFFF}0.2061 &
  {\color[HTML]{276221} \textbf{0.1334}} &
  \cellcolor[HTML]{FFFFFF}0.2104 &
  \cellcolor[HTML]{FFFFFF}0.1472 \\
\rowcolor[HTML]{EBF3FB}
DegMat-WLK     & 0.2510 & 0.2275 & 0.1814 & 0.1248 & 0.0419  & 0.0074  & 0.0643 & 0.0550  & 0.1580 & 0.0832 & 0.0000 & 0.0000 \\
\rowcolor[HTML]{FFFFFF}
Ecc-Smatch     & 0.1853 & 0.1311 & 0.2051 & 0.1023 & -0.1136 & -0.1452 & 0.0268 & -0.0176 & 0.1101 & 0.0073 & 0.1768 & 0.1075 \\
\rowcolor[HTML]{EBF3FB}
Ecc-S2match    & 0.1817 & 0.1222 & 0.2034 & 0.1011 & -0.1117 & -0.1477 & 0.0231 & -0.0251 & 0.1085 & 0.0035 & 0.1850 & 0.1102 \\
\rowcolor[HTML]{FFFFFF}
Ecc-SEMA       & 0.1800 & 0.1506 & 0.1471 & 0.0911 & -0.0207 & -0.0547 & 0.0271 & -0.0015 & 0.1041 & 0.0385 & 0.1542 & 0.1117 \\
\rowcolor[HTML]{EBF3FB}
Ecc-WLK        & 0.1821 & 0.1513 & 0.1601 & 0.0981 & -0.0089 & -0.0449 & 0.0168 & -0.0104 & 0.0824 & 0.0104 & 0.0000 & 0.0000
\end{tabular}%
}
\caption{
PRR at beam size $k=10$ for all uncertainty estimation methods across parsers and quality metrics (Smatch++ (S++) and SemBLEU (SB)). Higher PRR indicates better alignment between uncertainty estimates and prediction quality. Green cells denote the best-performing uncertainty estimator within each parser--metric pair. Consistent with the main findings, SPRING achieves the strongest overall reliability, while \textsc{DegMat-SEMA} frequently emerges as the most effective graph-based uncertainty estimator at larger beam sizes.
}
\label{tab:beam10_detailed}
\end{table*}

\begin{figure*}[!h]
\centering
\includegraphics[width=\textwidth]{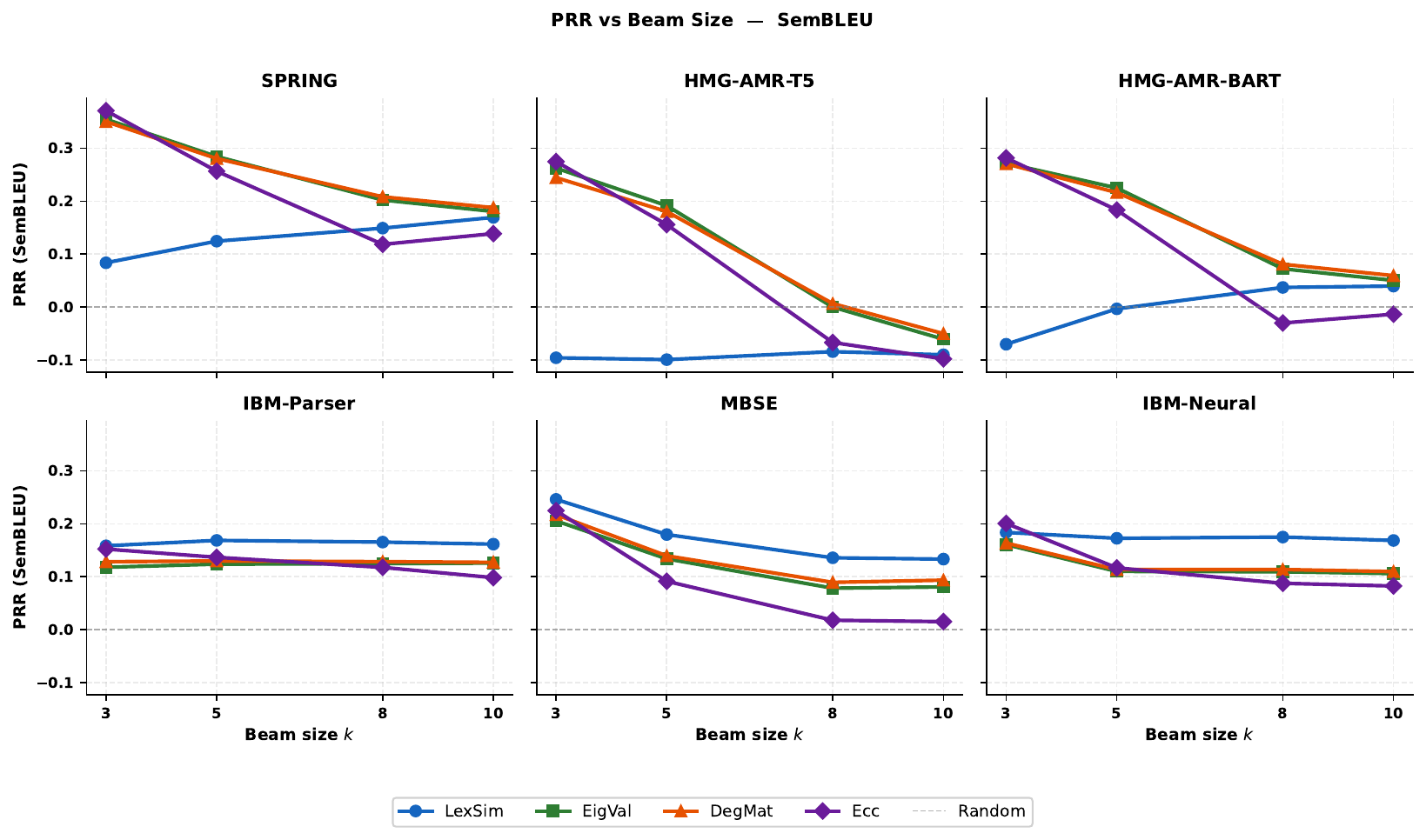}
\caption{
Parser-level PRR as a function of beam size evaluated using SemBLEU. Each subplot corresponds to one parser and compares uncertainty estimators across beam sizes.
}
\label{fig:prr_beam_sb}
\end{figure*}

\subsection{Relationship Between Parsing Performance and Reliability}
\label{amr_parser_smatch2}

Figure~\ref{fig:prr_vs_smatch} examines whether AMR parsing accuracy translates into more reliable uncertainty estimates. Each point represents a parser--beam combination (6 parsers $\times$ 4 beam sizes = 24 points per panel), where the x-axis shows the parser's fixed top-1 Smatch accuracy and the y-axis shows mean PRR averaged across all 13 uncertainty estimators for that parser and beam size. Because Smatch accuracy is fixed per parser, the four points per parser are stacked vertically, reflecting how PRR varies across beam sizes at the same accuracy level.

Across all evaluation metrics, a moderate positive relationship is observed ($\rho=0.388$ for Smatch++, $\rho=0.317$ for S$^2$Match, and $\rho=0.360$ for SemBLEU), indicating that parsers with stronger top-1 accuracy tend to achieve higher mean PRR across beam configurations. However, the correlations confirm that accuracy alone does not fully determine the quality of uncertainty. Parsers with comparable Smatch scores often occupy different vertical regions of the PRR space, suggesting that beam decoding behaviour and architectural design also independently influence reliability, beyond accuracy.

The vertical spread of points within each parser further illustrates the effect of beam size on uncertainty quality: some parsers maintain stable PRR across beams, while others experience substantial degradation as beam size increases, reinforcing the findings from RQ2. Finally, the variation in $\rho$ across evaluation metrics indicates that the strength of the performance--uncertainty relationship depends on how prediction quality is defined, consistent with the findings reported in RQ3.

Overall, parser accuracy provides only a partial explanation of uncertainty quality, and reliability emerges as a complementary evaluation dimension that is not captured by conventional AMR parsing metrics alone.

\begin{figure*}[!h]
\centering
\resizebox{\textwidth}{!}{
\includegraphics{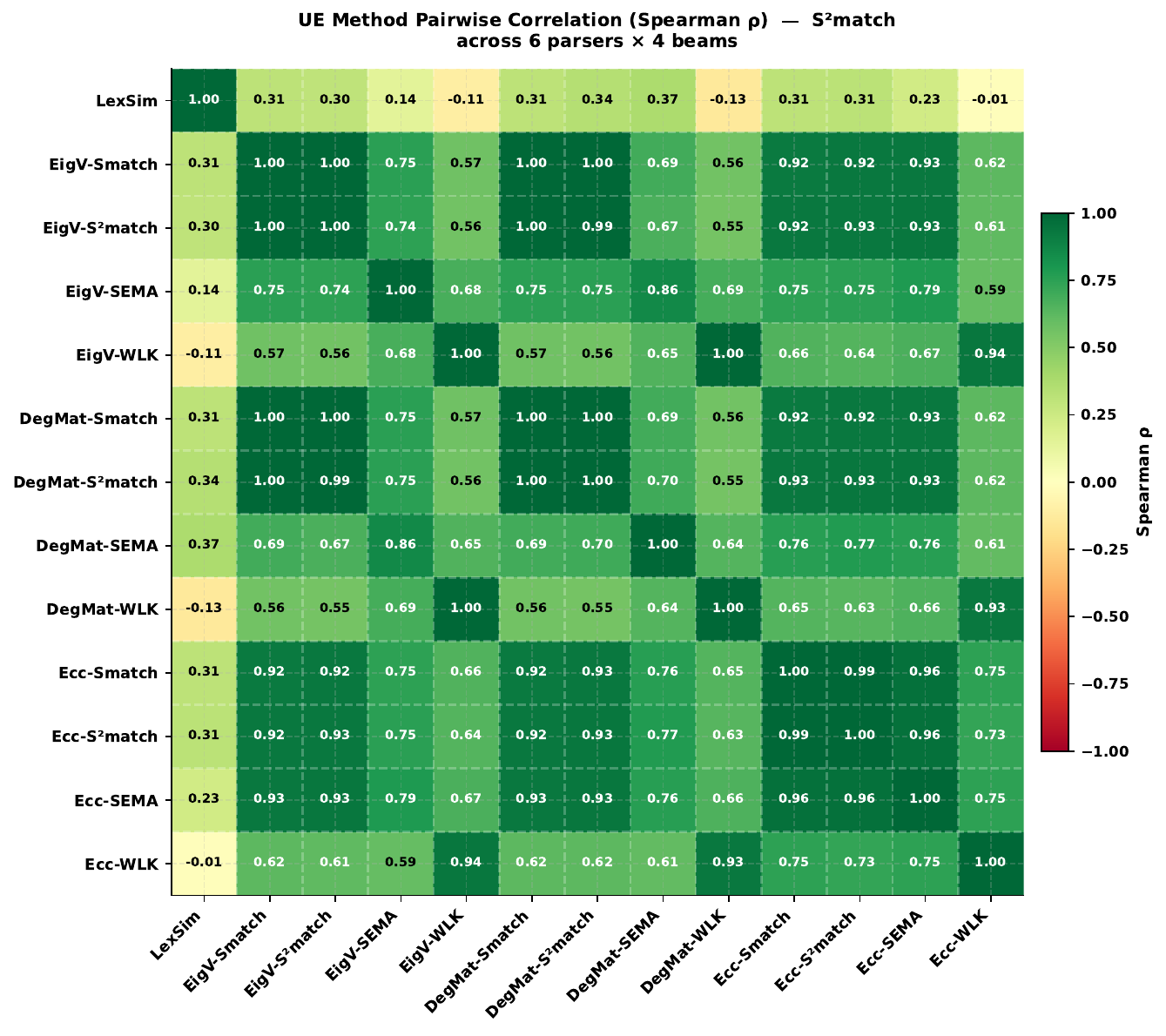}
}
\caption{
Pairwise Spearman correlation between uncertainty estimators evaluated using S$^2$Match.
}
\label{fig:ue_corr_s2}
\end{figure*}

\subsection{Uncertainty Estimator Stability Across Beam Sizes}
\label{amr_parser_RQ2}

Figure~\ref{fig:ue_stability} reports the mean PRR aggregated across all six parsers for each uncertainty estimator and beam size, directly supporting RQ2. The overall pattern is consistent with the main results: uncertainty-estimation reliability is highest at $k=3$ and generally declines as the beam size increases, with the sharpest drops occurring between $k=5$ and $k=8$.

Degradation rates differ markedly across estimators. \textsc{Ecc}-based methods start among the strongest at $k=3$, with \textsc{Ecc-S$^2$match} reaching 0.345 under S$^2$Match, yet show the steepest decline, falling to 0.141 at $k=10$. \textsc{EigV} variants follow a similar trajectory, dropping from approximately $0.30$--$0.34$ at $k=3$ to $0.14$ - $0.20$ at $k=10$, depending on the metric. Thus, both estimator families show substantial degradation in reliability as the beam size increases.

In contrast, \textsc{LexSim} is the only estimator whose mean PRR increases with beam size, rising from 0.149 to 0.171 under Smatch++ and from 0.187 to 0.215 under S$^2$Match, suggesting that its uncertainty estimates are less sensitive to increasing candidate diversity. Among graph-based methods, SEMA-based variants show greater robustness: \textsc{DegMat-SEMA} retains PRR values of 0.182 and 0.220 at $k=10$ under Smatch++ and S$^2$Match, respectively, substantially outperforming other graph-based methods at larger beam sizes.

Notably, S$^2$Match consistently yields higher mean PRR than Smatch++ across all estimators and beam sizes. For example, \textsc{EigV-S$^2$match} obtains 0.335 compared with 0.301 at $k=3$. This indicates that the observed effect of increasing beam size also depends on how prediction quality is measured, with S$^2$Match generally producing stronger uncertainty-reliability scores in this analysis. Overall, moderate beam sizes ($k=3$ to $k=5$) provide the strongest trade-off between candidate diversity and uncertainty-estimation reliability.

\begin{figure*}[!h]
\centering
\includegraphics[width=\textwidth]{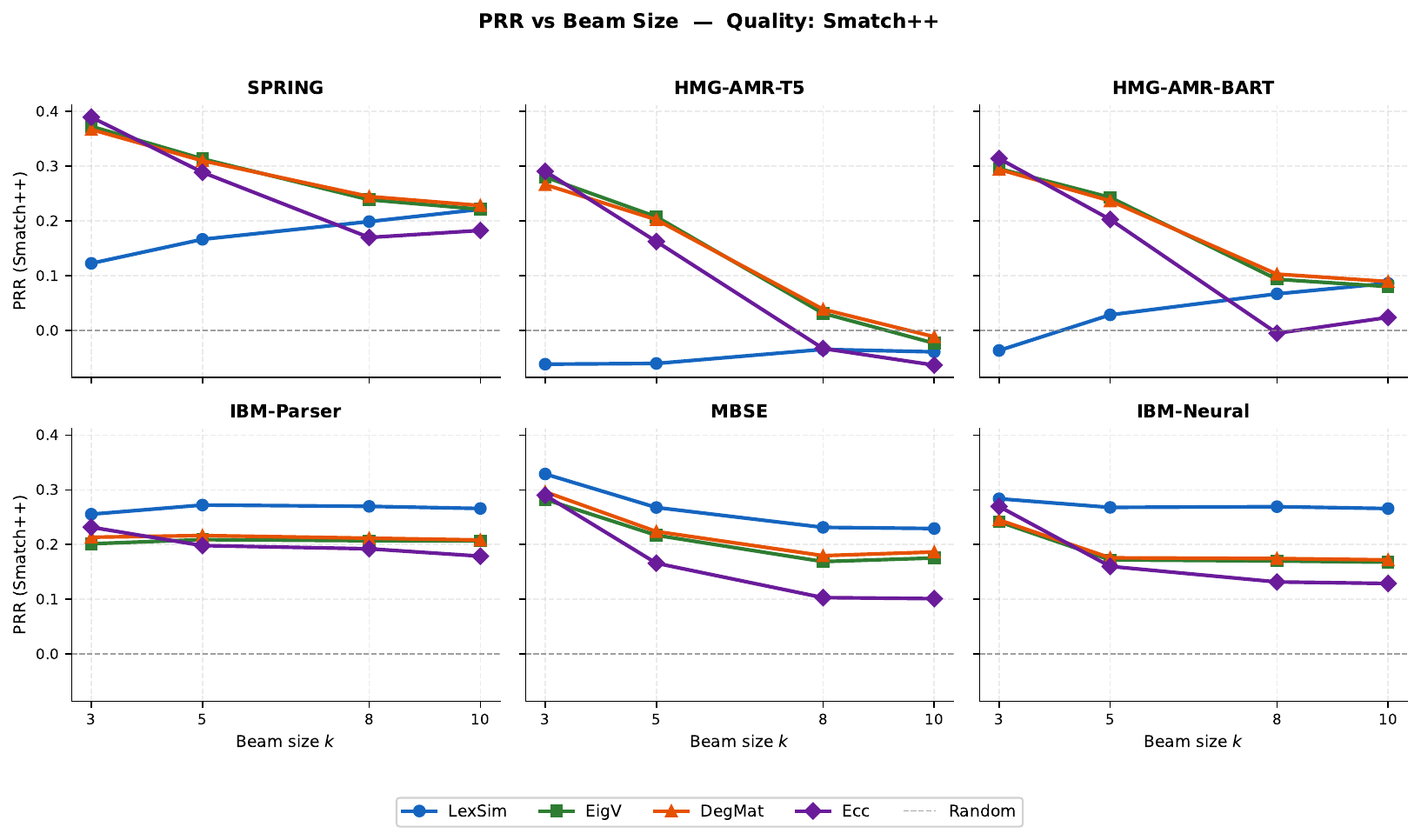}
\caption{
Parser-level PRR as a function of beam size evaluated using Smatch++. Each subplot corresponds to one parser and compares uncertainty estimators across beam sizes.
}
\label{fig:prr_beam_spp}
\end{figure*}

\begin{figure*}[!h]
\centering
\includegraphics[width=\textwidth]{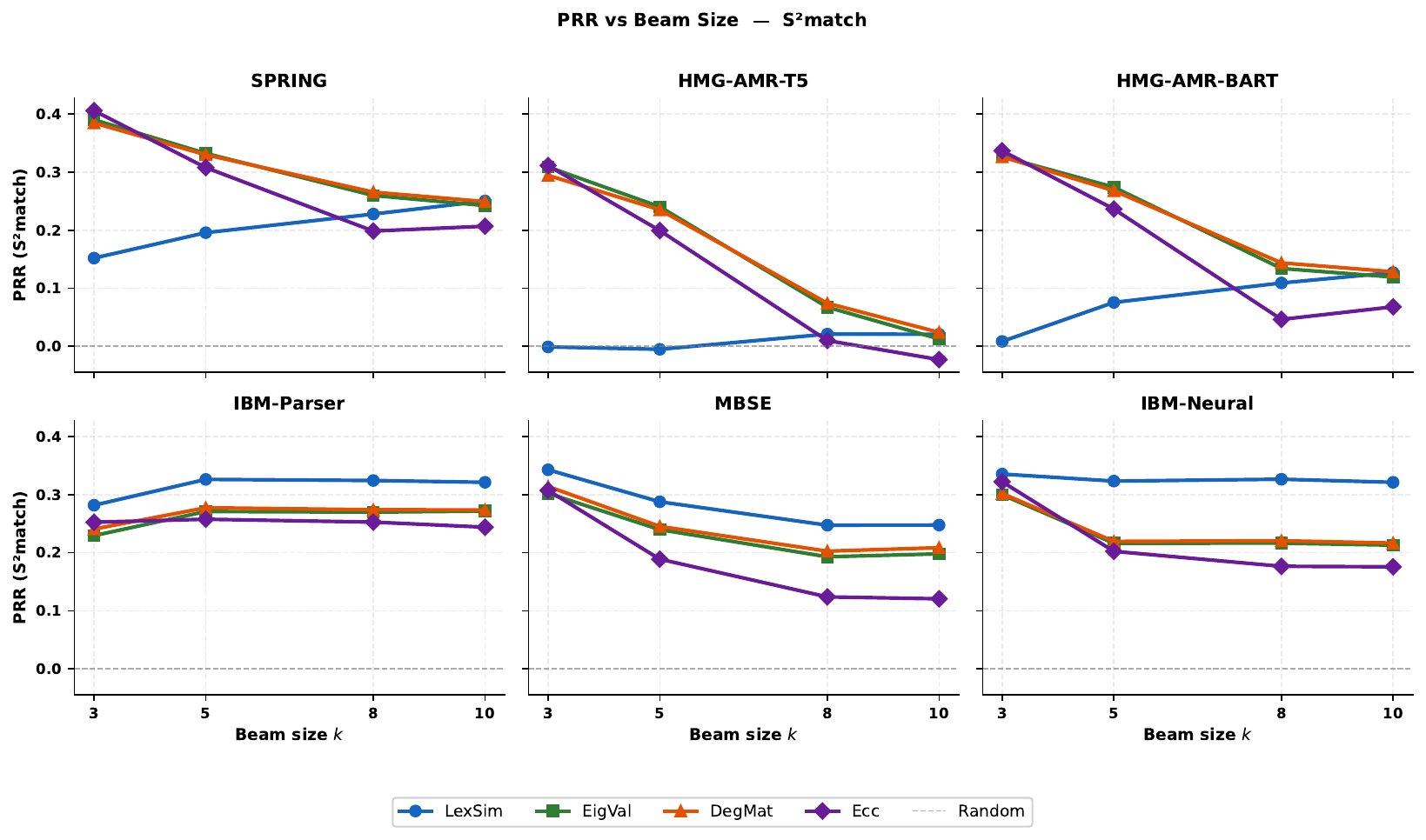}
\caption{
Parser-level PRR as a function of beam size evaluated using S$^2$Match. Each subplot corresponds to one parser and compares uncertainty estimators across beam sizes.
}
\label{fig:prr_beam_s2}
\end{figure*}

\begin{figure*}[!h]
\centering
\resizebox{\textwidth}{!}{
\includegraphics{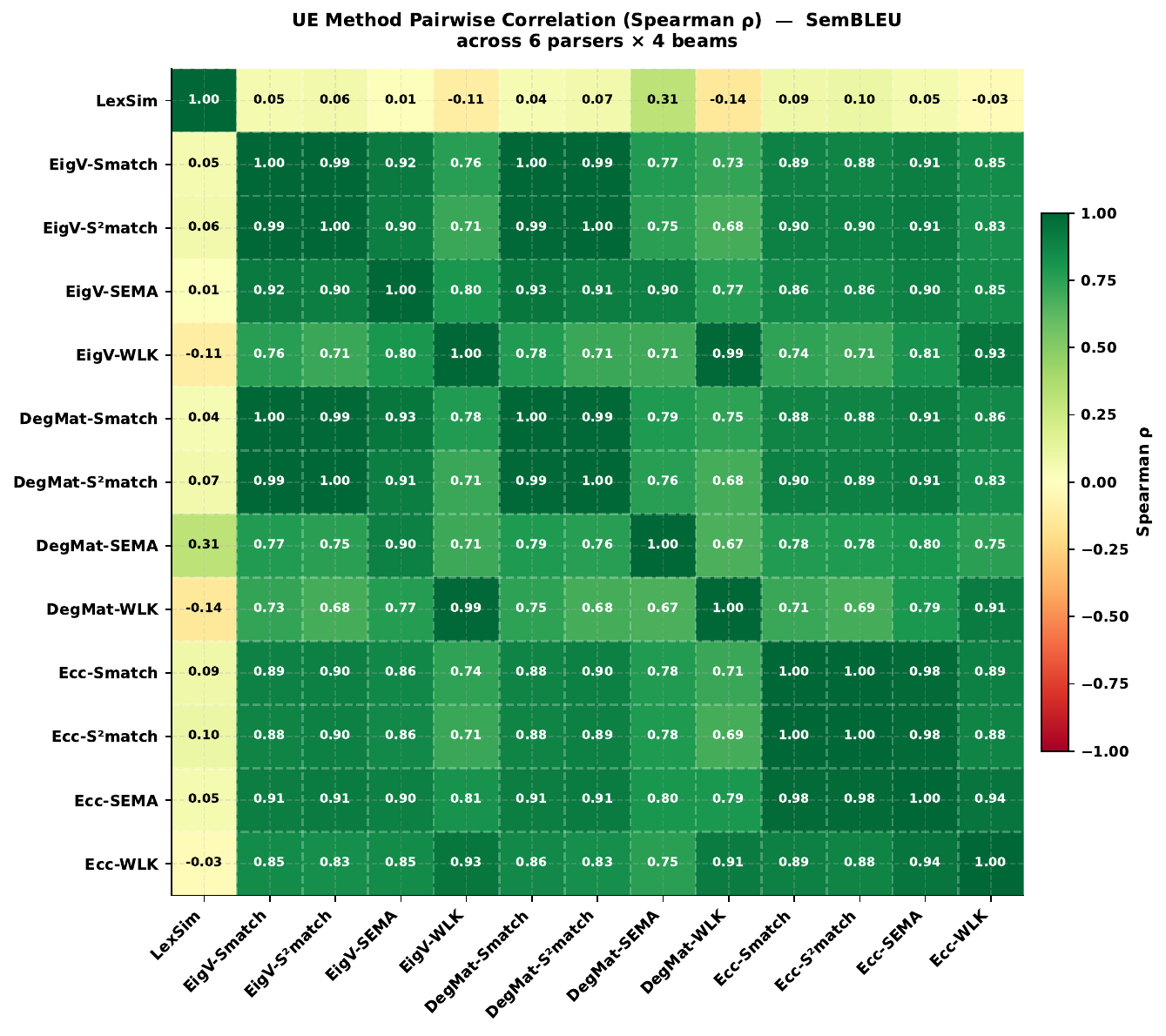}
}
\caption{
Pairwise Spearman correlation between uncertainty estimators evaluated using SemBLEU.
}
\label{fig:ue_corr_sb}
\end{figure*}

\subsection{Agreement Between UE}
\label{amr_parser_RQ3}

Figures~\ref{fig:ue_corr_spp},~\ref{fig:ue_corr_s2}, and~\ref{fig:ue_corr_sb} report pairwise Spearman correlations between uncertainty estimators across all parser and beam configurations. These results provide additional evidence for RQ3 by examining whether uncertainty estimators produce consistent rankings under different evaluation metrics.

Across all metrics, \textsc{EigV}, \textsc{DegMat}, and \textsc{Ecc} exhibit consistently strong mutual agreement. Under Smatch++, most pairwise correlations exceed $\rho=0.87$, with estimators using the same similarity function approaching near-perfect agreement (e.g.\ \textsc{EigV-Smatch} and \textsc{DegMat-Smatch}: $\rho=1.00$). Agreement is similarly high under S$^2$Match, where several pairs reach $\rho=1.00$, and remains strong under SemBLEU ($\rho\geq0.85$ for most graph-based pairs). This indicates that graph-based estimators generally preserve similar uncertainty rankings, although their agreement depends on the similarity function used.

In contrast, \textsc{LexSim} shows substantially weaker correlation with graph-based estimators across metrics. Under Smatch++, correlations range from $\rho=-0.16$ (\textsc{EigV-WLK}) to $\rho=0.31$ (\textsc{DegMat-SEMA}), while under SemBLEU they decrease to as low as $\rho=-0.14$ (\textsc{DegMat-WLK}). This indicates that lexical overlap captures an uncertainty signal that differs substantially from structural graph agreement, and that the degree of this difference varies with the evaluation metric.

Metric-dependent differences are also observed in the PRR results. Figure~\ref{fig:ue_stability} shows consistently higher mean PRR under S$^2$Match than Smatch++ across estimators and beam sizes; for example, \textsc{EigV-S$^2$match} obtains 0.335 compared with 0.301 at $k=3$. Under SemBLEU, WLK-based estimators show stronger agreement with other graph-based methods ($\rho\geq0.85$), reducing the separation between estimator families observed under Smatch++ and S$^2$Match. Overall, these results show that changing the evaluation metric can alter both uncertainty reliability and agreement between estimators, indicating that no single metric provides a complete view of uncertainty reliability.

\subsection{Parser Reliability Across Beam Sizes}
\label{amr_parser_RQ4_1}

Figures~\ref{fig:prr_beam_sb}, \ref{fig:prr_beam_spp}, and
\ref{fig:prr_beam_s2} visualise parser-level PRR across beam sizes, with each curve reporting the mean PRR for a given uncertainty estimator.

SPRING maintains the strongest and most consistent reliability profile across all metrics. Under Smatch++, PRR remains above $0.18$ at $k=10$ for all estimators except \textsc{Ecc}, which converges with others at larger beams. Under S$^2$Match and SemBLEU, SPRING similarly retains positive PRR across all estimators and beam sizes, confirming its robustness to candidate expansion.

IBM-Parser is the most stable parser across beam sizes. Under all three metrics, PRR remains nearly flat from $k=3$ to $k=10$, with
\textsc{LexSim} consistently outperforming graph-based methods and remaining the strongest estimator throughout. IBM-Neural shows a similar stability pattern, with \textsc{LexSim} again achieving the highest PRR at larger beams.

HMG-AMR-T5 exhibits the steepest degradation across all metrics. Under Smatch++, graph-based estimators drop below zero by $k=8$, and \textsc{LexSim} also approaches zero by $k=10$. Under SemBLEU, all estimators reach negative PRR at $k=8$ and $k=10$, indicating that beam candidates become actively misleading. HMG-AMR-BART shows similar sensitivity, particularly under SemBLEU, where \textsc{Ecc} drops below zero at $k=10$.

MBSE achieves a competitive PRR at $k=3$ under all metrics. However, it degrades noticeably at larger beams, particularly under S$^2$Match where \textsc{Ecc} falls to $\approx0.12$ at $k=10$, substantially below \textsc{LexSim} and \textsc{DegMat}. Overall, these results reinforce RQ4: the most reliable parser is the one that maintains reliable uncertainty estimates as candidate diversity increases not the one with the highest parsing accuracy.

\subsection{Detailed Uncertainty Estimation Results}
\label{amr_parser_RQ4_2}

Table~\ref{tab:beam10_detailed} reports complete PRR results at $k=10$ across all parsers, uncertainty estimators, and evaluation metrics. These results provide a more detailed view of RQ4 by examining whether the reliability of a parser is consistent across different uncertainty estimators at a large beam size. \textsc{DegMat-SEMA} achieves the strongest PRR for SPRING under both Smatch++ (0.271) and SemBLEU (0.247), and also leads for HMG-AMR-BART (0.137, 0.109) and HMG-AMR-T5 (0.069, 0.023). In contrast, \textsc{LexSim} achieves the strongest PRR for IBM-Parser (0.266, 0.162), IBM-Neural (0.266, 0.169), and MBSE (0.229, 0.133). Thus, the estimator providing the most reliable uncertainty signal at $k=10$ varies across parsers, indicating that parser comparison based on reliability is also influenced by the interaction between the parser and the uncertainty estimator.

\textsc{Ecc}-based methods perform weakest overall at $k=10$, with several parsers recording negative PRR under SemBLEU, such as HMG-AMR-BART ($-0.025$) and HMG-AMR-T5 ($-0.148$). This suggests that eigenspace-based candidate spread becomes less informative as beam size increases for these parsers. Overall, the detailed results show that although SPRING provides the strongest overall reliability, its advantage does not imply that a single uncertainty estimator is optimal across all parsers.

\end{document}